\documentclass[11pt]{article}
\usepackage{acl}

\usepackage{amsmath}
\usepackage{booktabs}
\usepackage{tabularx}
\usepackage{array}
\usepackage{makecell}
\usepackage{graphicx}
\usepackage{tabularx}
\usepackage{array} 
\usepackage{amssymb} 
\usepackage{booktabs}
\usepackage{siunitx}
\usepackage{rotating}  
\usepackage{dblfloatfix} 
\usepackage{placeins}   
\usepackage{dblfloatfix}   
\usepackage{placeins}      
\usepackage{graphicx} 
\usepackage[table]{xcolor} 
\definecolor{offgray}{gray}{0.45} 

\usepackage{booktabs}       
\usepackage{adjustbox}
\definecolor{coffee}{RGB}{111,78,55} 
\definecolor{rowgray}{gray}{0.92}
\definecolor{impgreen}{RGB}{0,140,0}
\usepackage{multirow}  
\usepackage{adjustbox}
\definecolor{coffee}{RGB}{111,78,55}
\colorlet{offgray}{black}
\usepackage{times}
\usepackage{latexsym}
\usepackage[T1]{fontenc}
\usepackage[utf8]{inputenc}
\usepackage{microtype}
\usepackage{inconsolata}
\usepackage{graphicx}
\title{StreamingEval: A Unified Evaluation Protocol towards Realistic Streaming Video Understanding}
\author{
Guowei Tang\textsuperscript{1},
Tianwen Qian\textsuperscript{1}$^{\dagger}$,
Huanran Zheng\textsuperscript{1},
Yifei Wang\textsuperscript{1},
Xiaoling Wang\textsuperscript{1} \\
\textsuperscript{1}East China Normal University, Shanghai, China \\
\small{\texttt{51274404096@stu.ecnu.edu.cn, twqian@cs.ecnu.edu.cn}}
}

\begin{document}
\maketitle
\begin{abstract}

Real-time, continuous understanding of visual signals is essential for real-world interactive AI applications, and poses a fundamental system-level challenge. Existing research on streaming video understanding, however, typically focuses on isolated aspects such as question-answering accuracy under limited visual context or improvements in encoding efficiency, while largely overlooking practical deployability under realistic resource constraints. To bridge this gap, we introduce StreamingEval, a unified evaluation framework for assessing the streaming video understanding capabilities of Video-LLMs under realistic constraints. StreamingEval benchmarks both mainstream offline models and recent online video models under a standardized protocol, explicitly characterizing the trade-off between efficiency, storage and accuracy. Specifically, we adopt a fixed-capacity memory bank to normalize accessible historical visual context, and jointly evaluate visual encoding efficiency, text decoding latency, and task performance to quantify overall system deployability. Extensive experiments across multiple datasets reveal substantial gaps between current Video-LLMs and the requirements of realistic streaming applications, providing a systematic basis for future research in this direction. Codes will be released at \url{https://github.com/wwgTang-111/StreamingEval1}.

\end{abstract}
\begingroup
\renewcommand{\thefootnote}{\fnsymbol{footnote}}
\footnotetext[2]{Corresponding author.}
\endgroup
\section{Introduction}

As video large language models (Video-LLMs)~\citep{maaz2024video_chatgpt,zhang2023video_llama,song2024moviechat,jin2024chat_univi} continue to advance, applications such as embodied robots~\citep{driess2023palme,brohan2023rt2, wang2026ocra}, live-streaming assistants~\citep{chen2024videollm_online,xu2025streamingvlm}, and autonomous driving systems~\citep{levinson2011towards, qian2024nuscenes, brodermann2025cafuser} are becoming increasingly feasible. In these scenarios, visual inputs arrive continuously in an incremental manner, requiring models to process them in real time under strict latency and resource constraints. However, existing mainstream Video-LLMs are designed and evaluated in offline settings, where the input videos are pre-recorded and fully accessible by the model. In contrast, streaming video understanding requires the model to process continuously arriving inputs without access to future frames, while sustaining instant responsiveness over potentially unbounded time horizons.

\begin{figure}[t]
  \centering
  \includegraphics[
    width=1.0\columnwidth,
    trim=7.5cm 1.6cm 7.5cm 1.6cm,
    clip
    ]{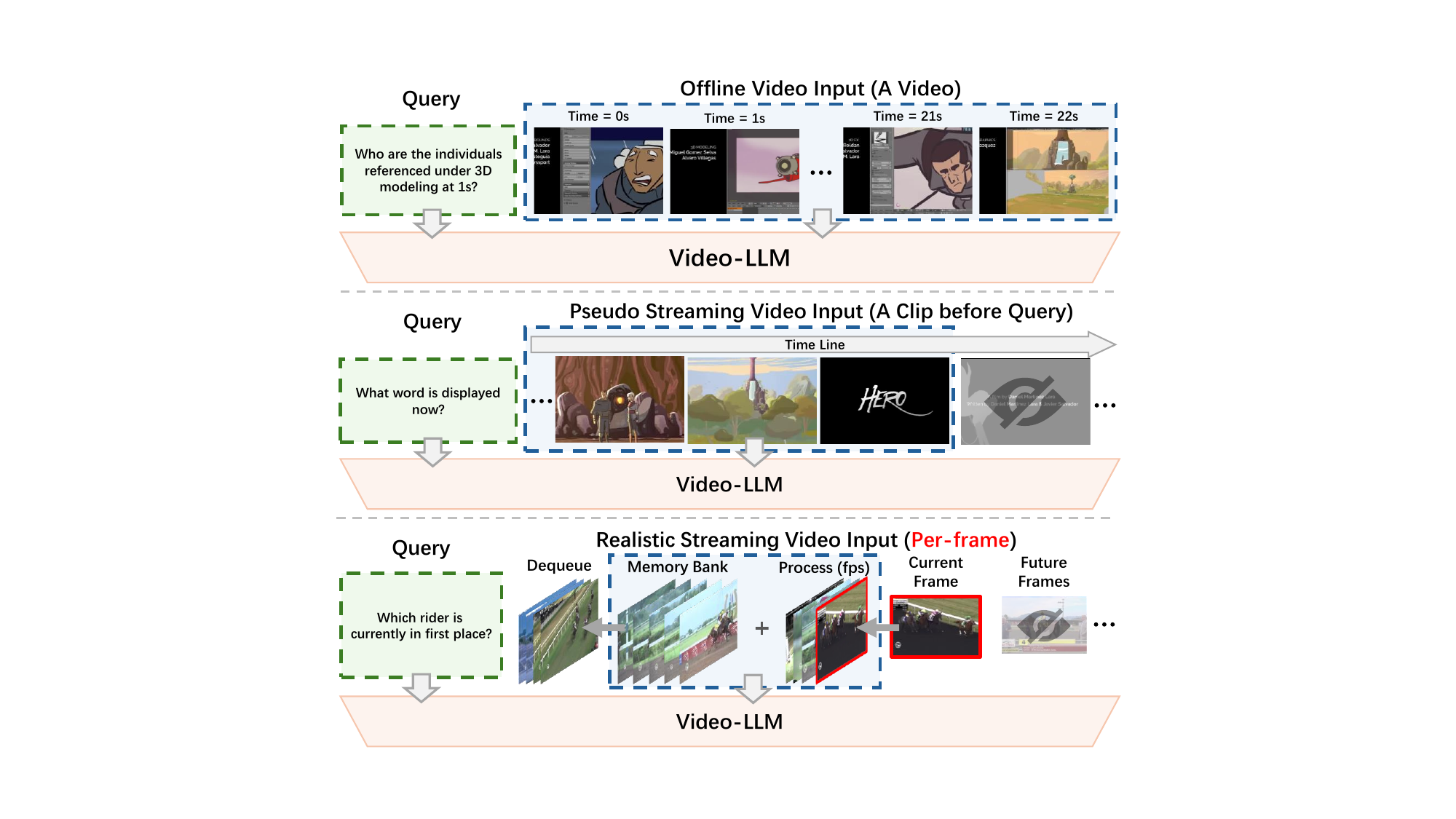} 
  \caption{Illustration of the conventional offline video understanding paradigm versus the streaming paradigm. Top: offline inference with full access to the video. Middle: pseudo-streaming inference, which truncates videos at query timestamps but still processes each clip in an offline manner. Bottom: realistic streaming inference with real-time incremental input and limited memory bank.}
  \label{fig:fig1}
\end{figure}

This fundamental mismatch between offline and streaming settings gives rise to several technical challenges that are largely absent in conventional video understanding tasks. First, streaming models must reason with incomplete and evolving visual context, as only past and current frames are available at any time. Second, streaming video is characterized by unbounded temporal duration, and long-running applications require persistent visual memory, but limited GPU resources inevitably lead to out-of-memory failures. Third, streaming systems impose stringent efficiency requirements. If visual encoding or memory updates fall behind the incoming frame rate, frames will accumulate and break real-time responsiveness. Similarly, excessive decoding latency directly degrades the interactive experience. Figure~\ref{fig:fig1} illustrates the differences between offline and streaming video understanding paradigms.

Recent studies have begun to investigate individual aspects of these challenges in isolation. For example, OVO-Bench~\citep{li2025ovobenchfarvideollmsrealworld} introduces an evaluation benchmark aligned with streaming context constraints and analyzes the model performance across different timestamps of questions (Backward Tracing, Real-Time Visual Perception, and Forward Active Responding). Several recent studies investigate the impact of visual encoding efficiency on online video processing~\citep{yao2025timechatonline,wang2025acceleratingstreamingvideollm,zeng2025streamforest,chen2025streamkvstreamingvideoquestionanswering,ning2025livevlm}, or quantify the effects of text decoding latency~\citep{reddi2019mlperf}. However, the diversity of evaluation methodologies has so far prevented the emergence of a unified and systematic evaluation framework. In particular, some benchmarks rely on pseudo-streaming settings, where videos are truncated at the query timestamp but still processed offline, limiting fair, reproducible, and deployment-oriented comparisons across models.

To bridge this gap, we introduce StreamingEval, a unified evaluation framework for video understanding under realistic streaming constraints. StreamingEval formalizes streaming video understanding as a system-level problem and provides a standardized evaluation protocol with metrics that jointly characterize accuracy, latency, and memory usage. Concretely, this framework implements a standardized streaming pipeline with three modules: (i) a frame player that continuously emits raw video frames at a fixed frame rate; (ii) an encoding-and-memory processor that performs per-frame visual encoding and memory updates according to each model’s design; (iii) and a response generator that, upon receiving a query, encodes the query, loads the current visual memory, and invokes the language model to generate answers. Based on this framework, we evaluate 12 representative online and offline Video-LLMs. Native online models retain their original streaming mechanisms and configurations, while offline models are modified with a fixed-capacity visual memory bank managed via a first-in-first-out (FIFO) policy, standardizing the accessible historical context during streaming inference.

Extensive experiments across multiple datasets demonstrate that current methods claimed to be ``online'' fail to operate reliably under strict streaming constraints. Moreover, under identical streaming settings, mainstream offline models often outperform specialized online models at the cost of higher resource consumption. Performance can be improved by enlarging memory capacity or raising visual input resolution, but this entails a trade-off with efficiency and deployability. Overall, our results reveal substantial gaps between current Video-LLMs and the requirements of realistic streaming applications, highlighting open challenges for future research. In summary, our main contributions are as follows:

\begin{itemize}
  \item We propose StreamingEval, a framework for assessing both the holistic capabilities of models and their deployability in streaming settings.
  \item We establish a scalable suite of online evaluation metrics and a unified protocol, enabling fair comparisons across models under consistent constraints.
  \item We conduct systematic empirical evaluations and analyses of representative state-of-the-art online and offline models, providing a reusable benchmark and clear directions for future research.
\end{itemize}

\section{Related Work}

\subsection{Video LLMs}

Offline and online VideoLLMs have emerged as two major paradigms in contemporary large-scale video model research~\cite{huang2025ovbench,chen2024videollm_online}. Offline approaches typically assume access to the entire video clip at inference time, and obtain video representations via strategies such as sparse sampling~\citep{feichtenhofer2019slowfast,bertasius2021timesformer}, global aggregation\citep{arnab2021vivit}, or long-video compression~\citep{ryoo2021tokenlearner}, which are then aligned with a language model for reasoning.~\cite{maaz2024video_chatgpt,zhang2023video_llama,lin2024video_llava,jin2024chat_univi,song2024moviechat,weng2024longvlm} Although these methods achieve strong performance on standard benchmarks, their underlying assumption—that the full video is available before inference—often deviates from real streaming applications. In contrast, online models are explicitly designed for streaming inputs and interactive use: they must incrementally encode newly arriving frames, continuously update memory/state, and respond immediately when a user query is issued. As deployment demands grow, research attention has increasingly shifted from optimizing offline accuracy alone to improving online usability and interactive experience\cite{chen2024videollm_online,huang2025ovbench,ning2025livevlm,wang2025acceleratingstreamingvideollm}.

\subsection{Streaming Benchmark}

Previous evaluations of streaming video understanding have largely emphasized answer accuracy. Early work such as VStream-QA\citep{flashvstream_llava}
 simulates realistic streaming queries by timestamping each question and restricting it to depend only on video content observed up to that moment, and it mainly measures correctness via accuracy and automatic scoring. StreamingBench\citep{lin2024streamingbenchassessinggapmllms}
 further broadens task coverage and modality dimensions, using overall performance as the primary metric to assess streaming understanding. OVOBench\citep{li2025ovobenchfarvideollmsrealworld} constructs multiple online VQA subtasks under three temporal contexts (past, present, and future).
 Recently, StreamEQA \citep{wang2025streameqa} establishes a streaming benchmark specifically tailored for embodied scenarios.
 However, in real-world online applications, latency is equally critical: it not only shapes user experience but is also tightly coupled with throughput, compute budgets, and serving costs. Reporting accuracy alone can therefore obscure a model’s practical usability under strict online constraints and the inherent accuracy–latency trade-off. To bridge this gap, our evaluation framework incorporates latency metrics alongside accuracy under a rigorous online setting, enabling reproducible and fair comparative analyses across streaming models.

\section{Streaming Evaluation Protocol}
In this subsection, we present StreamingEval in detail. We first provide a rigorous definition of streaming online inference, and then describe the concrete implementation pipeline of the evaluation framework. Next, we explain how to construct fair and reproducible comparison settings for both offline and online models. Finally, we introduce a set of evaluation metrics that are critical for online inference in streaming scenarios.

\subsection{Online Task Definition}

We define \emph{streaming online question answering} as a real-time, interactive multimodal dialogue setting in which the model continuously receives video frames along the temporal axis. At any time $t$, the input may consist of three components: (1) the interaction history between the user and the model up to time $t$, denoted as $C_t$; (2) the video frame acquired at time $t$, denoted as $V_t$; and (3) the user query issued at time $t$, denoted as $Q_t$. When the encoded query arrives at the model at time $t_1$, the backbone language model $f$ autoregressively generates a response $R_{t_1}$ conditioned on $\bigl(C_{t_1}, V_{\mathrm{enc}[0,t_1]}, Q_{t_1}\bigr)$, i.e., by maximizing the conditional probability $p\!\left(R_{t_1}\mid C_{t_1}, V_{[0,t_1]}, Q_{t_1}\right)$. Here, $V_{\mathrm{enc}[0,t_1]}$ denotes all video frames received up to time $t_1$ that have already been encoded by the model. This setting emphasizes the model's ability to perform inference and respond at arbitrary time points.

\subsection{StreamingEval Framework}

\begin{figure*}[t]
  \centering
 
  \includegraphics[
    width=0.8\textwidth,  
    trim=4.7cm 0.8cm 5.0cm 0.8cm,  
    clip
  ]{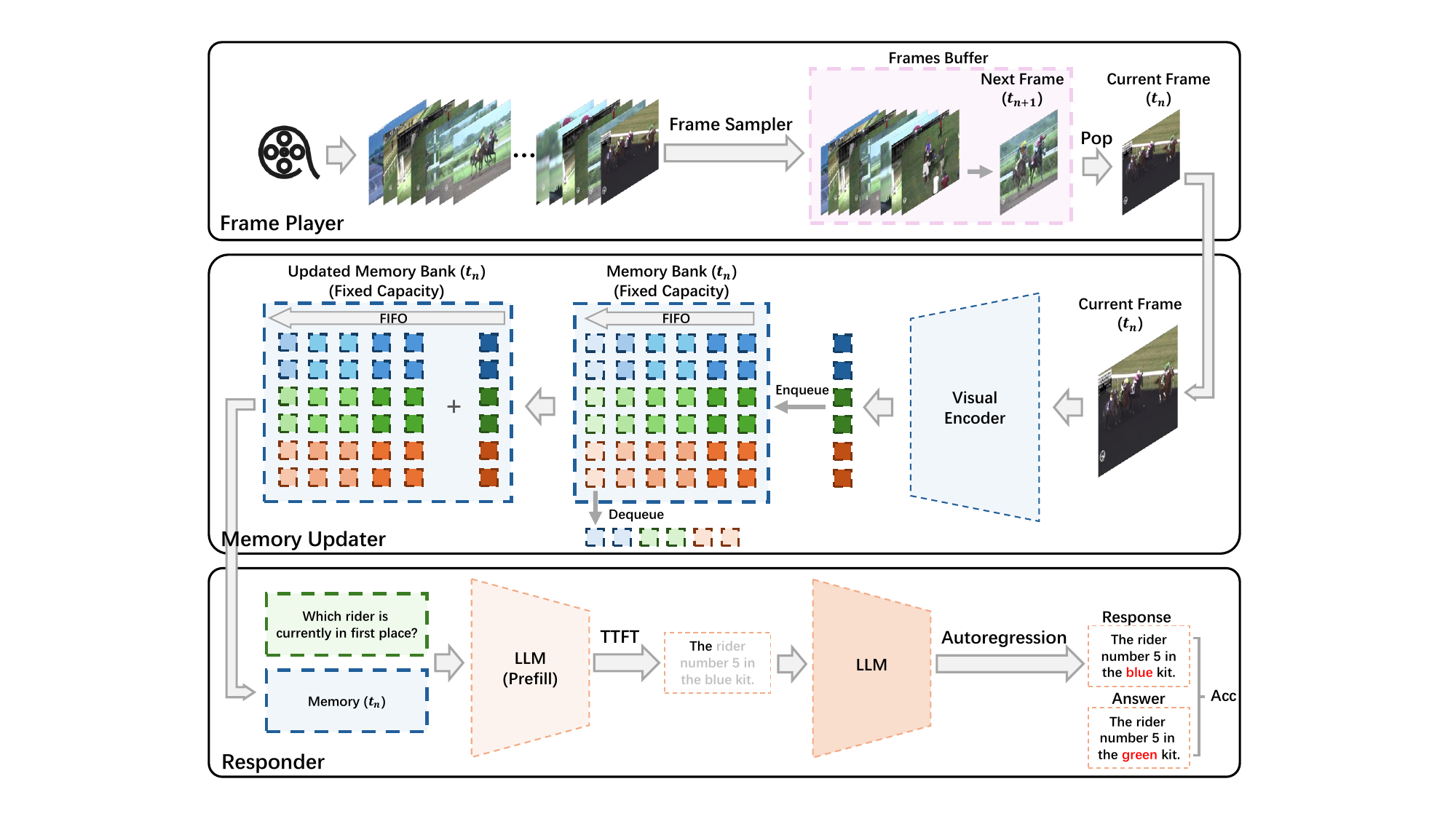}
  \caption{Overview of the StreamingEval framework. The framework standardizes streaming video understanding by modeling continuous input ingestion, incremental visual memory updates, and query-driven inference within a unified protocol.}
  \label{fig:fig2}
\end{figure*}

To enable streaming online inference, we implement the execution backend of StreamingEval as an asynchronous, time-causal pipeline composed of three decoupled processes that run in parallel, the framework is illustrated in Figure ~\ref{fig:fig2}. Specifically, it consists of a \textbf{Frame Player}, an \textbf{Encoder-and-Memory Updater}, and a \textbf{Responder}. These processes communicate via inter-process queues and/or shared buffers, emulating the behavior of an online system in which video frames arrive continuously, the model updates continually, and user queries may occur at any time, without introducing additional synchronization-induced blocking~\citep{ma2021simuleval}.

\paragraph{Input Stream and Frame Player.}
For an arbitrary video stream, we represent it as $\{(v_i,\tau_i)\}_{i=1}^{\infty}$, where $v_i$ denotes the $i$-th frame and $\tau_i$ denotes its arrival time. The frame player samples frames at a fixed interval $\rho$ and streams them to downstream processes.

\paragraph{Encoder \& Memory Updater.}
The encoder first maps each incoming frame to a visual representation:
\begin{equation}
\label{eq:zi}
z_i = g_\theta(v_i),
\end{equation}
where $g_\theta$ denotes the visual encoding backbone. The memory updater then maintains an online memory state $M$. For online models, the memory is updated according to the model-specific update rule $\mathcal{U}$ by writing the new representation into memory:
\begin{equation}
\label{eq:memory_update}
M_{\tau_i^+} = \mathcal{U}\!\left(M_{\tau_i^-}, z_i;\, B, \pi\right),
\end{equation}
where $\tau_i^-$ and $\tau_i^+$ denote the memory states immediately before and after the arrival of the $i$-th frame at time $\tau_i$, respectively. $B$ specifies the memory budget (capacity constraint), and $\pi$ denotes the corresponding write/eviction policy. For offline models, we apply a projection layer to map visual features into the embedding space aligned with the language model, and then store them in a fixed-length visual context window maintained with a first-in-first-out (FIFO) policy.

\paragraph{Responder.}
A user may launch a query $q_{t_0}$ at any time $t_0$. Once the query is triggered, the responder first encodes it, and denotes the time when encoding finishes as $t_1$. The responder then reads the memory snapshot available at time $t_1$, denoted by $M_{t_1}$, and conditions on the dialogue context $C_{t_1}$ together with the query $q_{t_0}$ to autoregressively generate an answer:
\begin{equation}
\label{eq:response_generation}
R_{t_1} \sim p_\phi(\cdot \mid q_{t_0}, C_{t_1}, M_{t_1}),
\end{equation}
where $C_{t_1}$ represents the interaction history up to time $t_1$, $M_{t_1}$ denotes the information updated up to time $t_1$, and $p_\phi$ is parameterized by the backbone language model.

\subsection{Comparable Online/Offline Setups}

In streaming settings, a fair comparison between native online models and general multimodal models is not straightforward: multimodal models typically assume access to the full video, whereas online models must operate under strict causal constraints and rely only on historical information that has already arrived and been processed. Moreover, visual token embeddings differ in dimensionality across models; therefore, constraining the history length solely by the \emph{number} of tokens can yield inconsistent actual memory footprints. To address this, StreamingEval adopts a ``two settings, one unified budget'' strategy: we preserve the native mechanisms of online models as much as possible, while introducing a resource-constrained offline adapter for multimodal models that constructs context under the same resource budget, enabling fair comparisons.

\paragraph{Native Online-model Setting.}

For native online models, we follow the original online mechanisms and configurations in their papers as closely as possible, including incremental encoding, memory/state updates, retrieval policies, and default input resolution and preprocessing. Models run within our multi-process emulator: frames arrive continuously at a fixed frame rate, the model updates its memory on the fly, and when a query is issued at time $t$, the responder can only access the memory snapshot that is \emph{available up to $t$} (i.e., processed before $t$). This ensures strict causality while staying faithful to each model's intended design.

\paragraph{Multimodal-model Adapter Setting.}
For multimodal models, we introduce a unified bounded-memory adapter. As video frames arrive, each model produces visual representations in its native manner; after a projection layer aligns them to the language model embedding space, the resulting representations are written into a fixed-capacity memory bank. When the memory bank exceeds the budget, we evict the oldest content using a deterministic FIFO policy. When a query is issued, we concatenate the current memory bank with the dialogue context as the model input, thereby simulating a deployable version of multimodal models under strict online constraints.

For offline models, we adopt a fixed-capacity memory bank with a FIFO eviction policy to enable as neutral and reproducible an evaluation as possible under realistic streaming constraints, while avoiding the additional algorithmic gains introduced by compression or summarization modules that could compromise fairness. At the same time, StreamingEval is open to different memory-management strategies: methods such as clustering-based compression, learned summarization, and KV compression can all be readily plugged in, and their effects will be naturally reflected in accuracy, latency, memory usage, and the final StreamingScore.

\paragraph{Unified Resource Budget.}
To avoid unfair comparisons where different models incur different GPU memory footprints under the same number of visual tokens due to embedding-dimensionality mismatch, we enforce a byte-level resource budget and cap the storage of historical context at $M$. We account only for the two components that scale with context length: the projected visual-token representations cached in the memory bank, and the language Transformer KV cache associated with these visual tokens for incremental inference.~\citep{kwon2023pagedattention,dao2022flashattention}Implementation details of the computations are provided in the appendix.

\subsection{Evaluation Metrics}

Online streaming multimodal question answering is simultaneously constrained by \emph{throughput, interactive latency, the effective historical-context budget, and task correctness}; no single metric suffices to characterize real-world deployability. Accordingly, we introduce the StreamingEval suite: we define four core metrics from four complementary system perspectives—encoding, decoding, memory, and task—and additionally report an overall score for \emph{summary reporting}, to more clearly reflect the trade-offs among accuracy, latency, throughput, and resource usage across methods.

\noindent\textbf{Visual Encoding:}
A streaming system must keep up with the incoming input rate; otherwise frame backlog accumulates and latencies quickly become unstable. Therefore, we use MaxFPS to measure the maximum input frame rate the model can \emph{sustain} without persistent backlog, capturing the real-time throughput ceiling of the visual encoding and memory-update pipeline.

\noindent\textbf{Text Decoding:}
User experience is largely determined by the latency of the response; therefore, we report \textbf{TTFT} (time-to-first-token), defined as the elapsed time from query arrival to the generation of the first token.~\citep{lperf_client_ttft,mlperf_loadgen_ttft_issue1593}

\noindent\textbf{Memory Overhead:}
A key constraint in streaming settings is how much historical context can be retained under limited resources, which directly impacts deployability and coverage; therefore, we use Memory\_bank to denote the budget of online-available visual history cache, aligning the effective amount of historical context that different models can leverage in streaming scenarios.

\noindent\textbf{Task Performance:}
The most straightforward measure of model performance is answer accuracy; therefore, we use question-answering accuracy to evaluate a model's capability on streaming online QA.

\noindent\textbf{Comprehensive
Metric:}
To more accurately quantify a model’s overall capability and its performance under streaming constraints, we integrate the metrics from the major contributing factors in streaming scenarios and define a composite metric,\emph{StreamingScore}:
\begin{equation}
\label{eq:streaming_score}
\mathrm{StreamingScore}(\mathbf{w})
\triangleq
\frac{\mathrm{MaxFPS}^{w_f}\cdot \mathrm{Acc}^{w_a}}
{\mathrm{TTFT}^{w_t}\cdot M^{w_r}},
\end{equation}
\begin{equation}
\label{eq:effective_M}
M \triangleq \mathrm{Mem}\cdot \ln(\mathrm{Params}),
\end{equation}
\begin{equation}
\label{eq:streaming_score_weights}
w_f,w_a,w_t,w_r \ge 0,\quad
w_f+w_a+w_t+w_r=1.
\end{equation}

This metric is designed to capture the trade-offs among accuracy, latency, throughput, and resource usage in streaming settings. A higher \emph{StreamingScore} indicates a model that maintains higher throughput while achieving better accuracy with lower time-to-first-token latency and lower resource consumption.

\section{Experiments}

We evaluate 12 mainstream multimodal and online models under the StreamingEval framework. We first describe the experimental setup and datasets, then report overall results, followed by analyses under different memory\_bank budgets and input image resolution ranges, and finally summarize the key findings.

\begin{table*}[t!]
\centering
\scriptsize
\setlength{\tabcolsep}{2.5pt}
\renewcommand{\arraystretch}{1.25}

\begin{adjustbox}{width=\textwidth,center}
\begin{tabular}{l c c c | c c c c c | c c}
\toprule
\multirow{2}{*}{\textbf{Model}} &
\multirow{2}{*}{\textbf{MaxFPS}} &
\multirow{2}{*}{\textbf{Mem(GB)}} &
\multirow{2}{*}{\textbf{TTFT}} &
\multicolumn{5}{c|}{\textbf{OVO-Bench}} &
\multicolumn{2}{c}{\textbf{StreamingBench}} \\
\cmidrule(lr){5-9}\cmidrule(lr){10-11}
& & & &
\textbf{Real-Time } &
\textbf{Backward } &
\textbf{Forward  } &
\textbf{Overall} &
\textbf{StreamingScore} &
\textbf{Real-time } &
\textbf{StreamingScore} \\
\midrule\midrule

\multicolumn{11}{c}{\cellcolor{coffee!20}\textbf{Open-source Offline VideoLLMs}} \\
\midrule
Qwen3-VL-8B~\citep{bai2025qwen3vl}                 & 8  & 0.5 & 0.20 & 78.73 & 51.82 & 43.46 & 58.00 & 2.21 & 77.31 & 2.37 \\
InternVL3.5-8B~\citep{wang2025internvl35}          & 7 & 0.5 & 0.21 & 74.31 & 40.89 & 46.14 & 53.78 & 2.04 & 77.96 & 2.24 \\
Llava-OV1.5-8B~\citep{an2025llavaonevision15} & 7  & 0.5 & 0.21 & 75.51 & 45.80 & 45.44 & 55.58 & 2.05 & 76.19 & 2.22 \\
MiniCPM-V4.5-8B~\citep{yu2025minicpmv45}           & 6  & 0.5 & 0.18 & 74.55 & 54.52 & 42.05 & 57.04 & 2.07 & 76.55 & 2.23 \\
VideoLLaMA3-7B~\citep{zhang2025videollama3}        & 7  & 0.5 & 0.56 & 55.73 & 45.32 & 45.88 & 48.44 & 1.58 & 68.90 & 1.72 \\
VideoChat-7B~\citep{li2023videochat}               & 9 & 0.5 & 0.52 & 69.77 & 40.25 & 43.27 & 49.95 & 1.73 & 72.22 & 1.89 \\

\midrule
\multicolumn{11}{c}{\cellcolor{coffee!20}\textbf{Open-source Online Video-LLMs}} \\
\midrule
Flash-VStream-7B~\citep{flashvstream_llava}        & 8  & 0.35 & 0.12 & 29.86 & 25.35 & 44.23 & 33.15 & 2.34 & 23.23 & 2.18 \\
Flash-VStream-7B*~\citep{flashvstream_qwen_repo}        & 1  & 0.66 & 1.31 & 59.88 & 46.43 & 47.41 & 50.31 & 0.74 & 74.48 & 0.81 \\
ReKV-7B~\citep{di2025rekv}                          & 5  & 0.50 & 1.29 & 57.34 & 44.69 & 44.35 & 48.00 & 1.18 & 64.53 & 1.27 \\
StreamForest-7B~\citep{zeng2025streamforest}       & 4  & 0.46 & 0.98 & 61.20 & 52.02 & 53.49 & 55.57 & 1.26 & 77.26 & 1.37 \\
TimeChat-Online-7B~\citep{yao2025timechatonline}   & 7  & 0.34 & 0.62 & 58.60 & 42.00 & 36.40 & 45.67 & 1.67 & 73.64 & 1.88 \\
VideoChatOnline-4B~\citep{huang2025ovbench}        & 0.14 & 0.32 & 0.18 & 43.97 & 37.40 & 39.69 & 40.40 & 0.92 & 58.81 & 1.19 \\

\bottomrule
\end{tabular}
\end{adjustbox}

\caption{Overall results on OVO-Bench and StreamingBench. For OVO-Bench, we report the average scores of the Real-Time, Backward, and Forward task clusters; for StreamingBench, we report the overall Real-time average score, where the streaming score is computed following the definition described in the main text. Detailed results are provided in Table \ref{tab3}.}

\label{tab:overall}
\end{table*}

\subsection{Experiment Settings}

\label{sec:exp_setup}

We evaluate representative multimodal models and online video models within the StreamingEval framework. All experiments are conducted on a single RTX 4090 (48GB) GPU with BF16 inference, using a unified prompt and decoding configuration. TTFT is measured in wall-clock time. Streaming inputs are fed at 1fps. Due to space limitations, more experimental setups are shown in the Appendix \ref{app2}.

\subsection{Datasets}
\label{sec:datasets}
We evaluate StreamingEval on two widely used benchmarks for online/streaming video understanding: OVO-Bench \cite{huang2025ovbench} and StreamingBench \cite{lin2024streamingbenchassessinggapmllms}. Both benchmarks cover diverse video understanding and question answering formats, designed to test temporal reasoning, event recognition, and multimodal inference.

\paragraph{OVO-Bench.}
OVO-Bench is designed to evaluate online video understanding capabilities and assess models in three typical online scenarios: Backward Tracing, Real-Time Understanding, and Forward Active Responding. It comprises 12 tasks and approximately 2,800 fine-grained annotations with precise timestamps, and supports systematic query-based evaluation along the video timeline.

\paragraph{StreamingBench.}
StreamingBench organizes its task taxonomy around three dimensions: real-time visual understanding, multi-source (audio–visual) understanding, and contextual understanding. The benchmark comprises 18 tasks, 900 video clips, and 4,500 human-annotated QA pairs, and issues multiple queries at different timestamps within the same video to simulate continuous streaming inputs.

\subsection{Main Results}

\label{sec:overall_results}
Table~\ref{tab:overall} summarizes the results of 12 mainstream multimodal and online models under StreamingEval. We report task-level performance , system-level deployability metrics , and provide a normalized composite score, StreamingScore, for convenient comparison.

\paragraph{Encoding Efficiency.}
Most models achieve a peak throughput (MaxFPS) that satisfies the minimum requirement for streaming online inference (1 FPS), indicating basic real-time feasibility on the input side. VideoChatOnline is a notable exception: its MaxFPS is substantially below 1 FPS, revealing a severe bottleneck in the video encoding/state-update pipeline. We attribute this to its relatively complex memory-update and cross-frame maintenance mechanism, which significantly reduces the number of frames that can be processed per unit time. At such throughput, the system cannot keep pace with the incoming video stream. In practice, this would induce persistent backlog and accumulating latency, rendering the model difficult to deploy in real-world streaming settings.

\paragraph{Time-to-First-Token.}

For most models, the TTFT is primarily governed by the size of the visual-memory context window and the overhead of cross-modal context construction, yet it remains below 1.5 seconds in nearly all cases. Consequently, assuming the encoding throughput is sufficient, this level of decoding startup latency is unlikely to constitute a primary obstacle to practical deployment.

\paragraph{Overall Capability Comparison: Offline vs.\ Online Models}

As shown in Table~\ref{tab:overall}, under our strict streaming protocol, offline VideoLLMs and native online models exhibit a relatively consistent gap in question-answering accuracy. Overall, offline models tend to achieve higher accuracy on the aggregate scores of OVO-Bench and StreamingBench, reflecting their stronger representation and reasoning capacity, as well as their advantage in more fully leveraging visual evidence during answer generation. In contrast, to satisfy causal constraints and limited state updates, native online models typically rely on incremental memory mechanisms that compress, truncate, or selectively retrieve historical information, which is more prone to information loss on tasks requiring long-range temporal dependencies or fine-grained cues, thereby resulting in lower accuracy. Moreover, from the perspective of streaming interactivity, although offline models are often superior in accuracy, their streaming scores on OVO/SB do not necessarily lead accordingly; instead, some native online models attain higher streaming scores by offering faster responses and a more stable online pace under causal constraints, further revealing a systematic trade-off between strict streaming deployability and task accuracy. Overall, the results indicate a practical tension between strict online deployability and task accuracy: retaining critical information more effectively under limited resources remains a key direction for improving accuracy in streaming scenarios.

\paragraph{Performance Differences Across Task Categories.}
The per-category results in Table~\ref{tab:overall} reveal systematic performance divergences between offline and online VideoLLMs across task types.

For Backward Tracing, which relies on long-horizon temporal integration and reasoning over extended visual history, offline models are typically stronger, suggesting that full-context modeling better supports retrospective inference and evidence aggregation.

By contrast, for Forward Active Responding---which emphasizes rapid decision making and incremental interaction under strict streaming constraints---online models are largely on par with offline counterparts, and can even outperform them in some cases, highlighting the structural benefits of native online mechanisms for continuous updates and timely responses.

For Real-Time Visual Perception, offline models remain generally superior overall. This indicates that even when the task is more ``current-segment'' oriented, online approaches may still suffer from information compression and noise accumulation induced by limited buffering and incremental updates, which can undermine the stability of fine-grained visual representations compared to offline models with unified context encoding.

\subsection{Further Discussion}

\begin{figure}[t]
  \centering
  \includegraphics[width=\columnwidth,height=0.82\textheight,keepaspectratio]{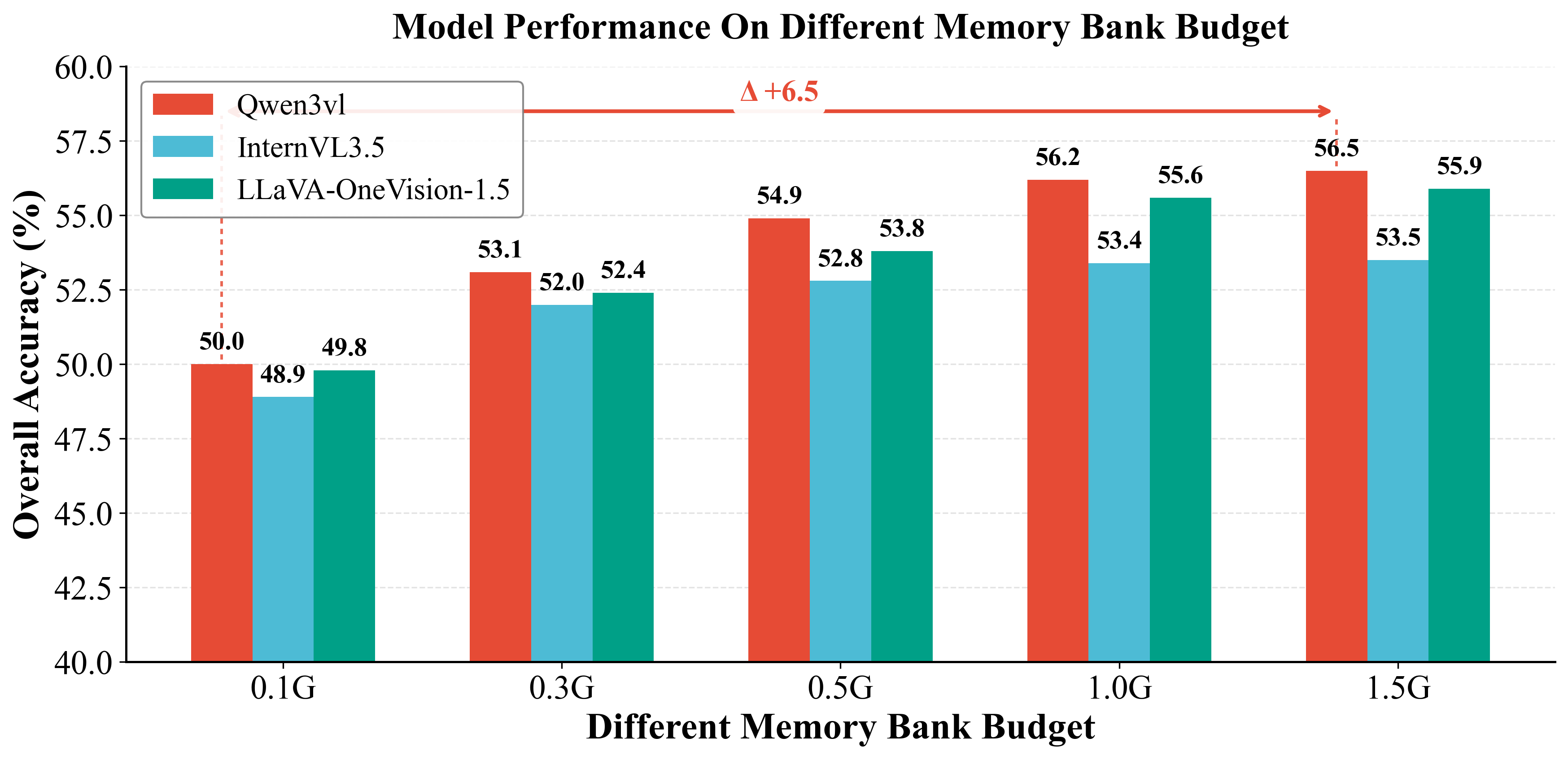}
\caption{Overall accuracy versus memory\_bank budget for three representative models. Detailed results are provided in Table \ref{tab:ovo_detail}.}

  \label{fig2:experiment}
\end{figure}

\paragraph{Sensitivity to the Memory\_bank Budget.}
\label{sec:memory_sensitivity}
To evaluate model performance under different historical-cache budgets, we keep all other protocol settings fixed and vary the Memory\_bank budget over \{1.5G, 1.0G, 0.5G, 0.3G, 0.1G\}, reporting the resulting changes in Overall Accuracy (Figure ~\ref{fig2:experiment}). As shown, accuracy nearly saturates in the relatively ample regime of 1.0G--1.5G: increasing the budget from 1.0G to 1.5G yields only marginal gains, indicating diminishing returns from further cache expansion. In contrast, as the budget is tightened from 1.0G to 0.5G, 0.3G, and finally 0.1G, all three models exhibit a clear accuracy drop, suggesting that performance is largely constrained by a historical-information bottleneck. While the overall ranking remains consistent, the scores converge substantially under the most restrictive 0.1G setting, implying that different methods are more strongly capped by the shared cache constraint. Meanwhile, Qwen3-VL still retains relatively higher accuracy at this extreme budget, reflecting more efficient extraction and utilization of limited historical information and better suitability for resource-constrained deployment.

\paragraph{Different Input Resolution.}

\label{sec:resolution_and_discussion}

Different input resolution determines how much visual detail is available per frame. With \texttt{Memory\_bank} fixed (set to the default in Table~\ref{tab:resolution_ablation}) and all other protocol settings unchanged, we evaluate three common input sizes (448$\times$448, 336$\times$336, and 224$\times$224), using a unified video-frame resizing pipeline to ensure consistent processing (Table~\ref{tab:resolution_ablation}). As shown in Table~\ref{tab:resolution_ablation}, all three representative models achieve their best accuracy at the highest resolution of 448$\times$448; as the resolution decreases, performance drops to varying degrees, indicating a stable degradation when downscaling inputs. The models also differ in robustness to resolution reduction: Qwen3-VL degrades more mildly, whereas InternVL~3.5 shows a more pronounced drop, suggesting stronger reliance on the additional fine-grained details provided by higher resolution; Flash-VStream also deteriorates as resolution decreases and remains substantially behind the other two in absolute accuracy. From the perspective of ``downscaling loss,'' the drop from 448$\times$448 to 336$\times$336 is relatively small, while the drop from 336$\times$336 to 224$\times$224 is more evident, implying that further downscaling more quickly hits a detail bottleneck. Overall, higher resolution improves accuracy but comes with higher computation and latency costs; under deployment budget constraints, 336$\times$336 is often a more robust trade-off between effectiveness and cost, whereas 224$\times$224 is cheaper but incurs a notably larger performance decrease.

\begin{table}[t]
\centering
\small
\setlength{\tabcolsep}{4pt}
\renewcommand{\arraystretch}{1.15}
\begin{tabular}{@{}l S[table-format=2.2] S[table-format=2.2] S[table-format=2.2]@{}}
\toprule
\textbf{Input Res.} &
\multicolumn{1}{c}{\textbf{Qwen3-VL}} &
\multicolumn{1}{c}{\textbf{InternVL 3.5}} &
\multicolumn{1}{c}{\textbf{FlashVstream}} \\
\midrule
224$\times$224 & 51.99 & 48.93 & 33.15 \\
336$\times$336 & 52.95 & 51.70 & 34.67 \\
448$\times$448 & 53.34 & 52.75 & 35.73 \\
\bottomrule
\end{tabular}
\caption{Overall results under different input resolutions on OVO-Bench. We report the overall score for each model under 224$\times$224, 336$\times$336, and 448$\times$448 inputs. Detailed results are provided in Table \ref{tab6}.}

\label{tab:resolution_ablation}
\end{table}

\subsection{Outlook and Future Directions}
Based on a systematic experimental analysis using StreamingEval, we observe that in online video understanding settings, a model’s offline accuracy does not necessarily translate into a usable online experience. Our evaluation indicates that different models exhibit significant and stable trade-offs along four dimensions-accuracy, throughput, latency, and limited history. In practice, the key bottlenecks that shape user experience often stem from response waiting time, performance degradation under throughput constraints, and information loss and error accumulation induced by restricted historical context.

Accordingly, we argue that future research on online video understanding models should more explicitly incorporate system constraints into model design objectives. First, at the architectural level, priority should be given to representation and memory mechanisms with streaming, incremental update capabilities, enabling models to reliably maintain critical states under limited history. Second, at the inference level, it is important to explore latency-accuracy controllable computation allocation strategies, so as to provide effective responses for real-time interactive use.

\section{Conclusions}
\label{sec:conclusion}
We propose StreamingEval, a unified evaluation protocol for video question answering models under strictly streaming, online constraints. Beyond accuracy, we emphasize deployability-oriented metrics, including time to first token (TTFT), MaxFPS, and a constrained memory bank, and introduce a resource-budget adapter to enable fair comparisons between online models and offline VideoLLMs. Using this protocol, we evaluate 12 representative models and show that being “online” does not necessarily translate into practical deployability; moreover, changes in the memory bank and input resolution lead to corresponding shifts in accuracy. We hope StreamingEval will facilitate reproducible, system-level evaluation and inspire future work to optimize for the end-to-end online user experience.

\section{Limitations}

Due to the lack of publicly available code or reproducible interfaces for some closed-source/proprietary models, we cannot reliably standardize their inference configurations or faithfully measure their true performance under a unified evaluation protocol and runtime environment. Consequently, our evaluation does not provide a comprehensive comparison across all mainstream closed-source models.
Limited computational resources and our focus on mobile/edge deployment constrain our experiments mainly to 7B/–8B scale models. As a result, we are unable to draw firm conclusions about the performance–efficiency trade-offs of larger models under the same streaming setting.
In future work, we will seek additional GPU resources and expand our reproducible coverage to evaluate larger models. We will also continuously update the project website with new results and analyses.

\paragraph{Potential Risks}
This work proposes an evaluation protocol only and does not include or release any dataset; therefore, the privacy and licensing risks are limited.
The main risks are that the protocol may be over-generalized as a universal standard, or may incentivize over-optimization for the reported metrics that does not reflect real-world performance.
We recommend validating the protocol across diverse domains/datasets/scenarios with robustness and subgroup analyses, and caution against using a single evaluation result to justify deployment or decision-making in high-stakes settings.

\section*{Acknowledgements}
Project supported by Shanghai Municipal Science and Technology Major Project 2025SHZDZX025G16.

\bibliography{custom}
\clearpage
\appendix
\section{Appendix}
We include supplementary material in the appendix to facilitate a deeper understanding of our experimental setup and results. Due to space constraints, we focus on the most essential details in each section. The appendix follows the structure of the main paper:Appendix I elaborates on the evaluation protocol, Appendix II provides additional experimental details.

\subsection{Detailed Evaluation Protocol}

For the $i$-th model, let $B$ denote the number of visual tokens retained. We define the linearly growing storage cost as
\[
\mathrm{Mem}_i(B)
= B\cdot d_i \cdot s_{\mathrm{emb}}
\;+\;
B\cdot 2L_i \cdot h^{\mathrm{kv}}_i \cdot s_{\mathrm{kv}},
\]
where $d_i$ is the embedding dimension of the projected visual tokens and $s_{\mathrm{emb}}$ is the bytes per element for these embeddings; $L_i$ is the number of layers in the language backbone;
$h^{\mathrm{kv}}_i$ is the per-layer KV channel width (e.g., $h^{\mathrm{kv}}_i = n^{\mathrm{kv}}_i \cdot d^{\mathrm{head}}_i$, with $n^{\mathrm{kv}}_i$ the number of KV heads and $d^{\mathrm{head}}_i$ the per-head dimension).
The factor $2$ accounts for both Key and Value tensors, and $s_{\mathrm{kv}}$ is the bytes per element for KV cache (e.g., $s_{\mathrm{kv}}{=}2$ for BF16).
Accordingly, we convert the byte budget $M$ (denoting $M_{\text{bytes}} = M \cdot 10^9$) into the model-specific visual-token cap:
\[
B_i=\left\lfloor \frac{M_{\text{bytes}}}{d_i s_{\text{emb}} + 2L_i h_i^{kv} s_{kv}} \right\rfloor .
\]

\subsection{Experimental Details}
\label{app2}
In this section, we first describe the experimental setup and environment. We then present the measurement results for each subtask in Section~4.3, followed by the experimental results discussed in Section~4.4.

\begin{table*}[t]
\centering
\scriptsize
\setlength{\tabcolsep}{3.2pt}
\renewcommand{\arraystretch}{1.15}
\begin{adjustbox}{max width=\textwidth,center}
\begin{tabular}{l|rrrrrrr|rrrr|rrrr|rrrrrrrrrrr}
\toprule
\multirow{3}{*}{\textbf{Model}} &
\multicolumn{15}{c|}{\textbf{OVO-Bench}} &
\multicolumn{11}{c}{\textbf{StreamingBench}} \\
\cmidrule(lr){2-16}\cmidrule(lr){17-27}
& \multicolumn{7}{c|}{\textbf{Real-Time Visual Perception}} &
  \multicolumn{4}{c|}{\textbf{Backward Tracing}} &
  \multicolumn{4}{c|}{\textbf{Forward Active Responding}} &
  \multicolumn{11}{c}{\textbf{Real-time Visual Understanding}} \\
\cmidrule(lr){2-8}\cmidrule(lr){9-12}\cmidrule(lr){13-16}\cmidrule(lr){17-27}
& OCR & ACR & ATR & STU & FPD & OJR & Avg. &
  EPM & ASI & HLD & Avg. &
  REC & SSR & CRR & Avg. &
  CS & OP & ATP & PR & ACP & SU & EU & CT & TR & CR & Avg. \\
\midrule\midrule

\multicolumn{27}{c}{\cellcolor{coffee!20}\textbf{Open-source Offline VideoLLMs}} \\
\midrule
\textcolor{offgray}{Qwen3-VL-7B} &
\textcolor{offgray}{91.95} & \textcolor{offgray}{78.90} & \textcolor{offgray}{79.31} & \textcolor{offgray}{67.98} & \textcolor{offgray}{73.27} & \textcolor{offgray}{80.98} & \textcolor{offgray}{78.73} &
\textcolor{offgray}{48.82} & \textcolor{offgray}{61.49} & \textcolor{offgray}{48.92} & \textcolor{offgray}{51.82} &
\textcolor{offgray}{21.78} & \textcolor{offgray}{64.86} & \textcolor{offgray}{50.42} & \textcolor{offgray}{43.46} &
\textcolor{offgray}{89.27} & \textcolor{offgray}{81.74} & \textcolor{offgray}{74.79} & \textcolor{offgray}{75.00} & \textcolor{offgray}{88.78} & \textcolor{offgray}{72.36} &
\textcolor{offgray}{68.55} & \textcolor{offgray}{40.41} & \textcolor{offgray}{85.67} & \textcolor{offgray}{71.88} & \textcolor{offgray}{77.31} \\
\textcolor{offgray}{InternVL3.5-8B} &
\textcolor{offgray}{88.59} & \textcolor{offgray}{75.23} & \textcolor{offgray}{77.59} & \textcolor{offgray}{61.80} & \textcolor{offgray}{71.29} & \textcolor{offgray}{73.91} & \textcolor{offgray}{74.31} &
\textcolor{offgray}{51.18} & \textcolor{offgray}{56.75} & \textcolor{offgray}{11.83} & \textcolor{offgray}{40.89} &
\textcolor{offgray}{25.36} & \textcolor{offgray}{67.73} & \textcolor{offgray}{50.00} & \textcolor{offgray}{46.14} &
\textcolor{offgray}{89.59} & \textcolor{offgray}{81.74} & \textcolor{offgray}{76.49} & \textcolor{offgray}{78.70} & \textcolor{offgray}{88.78} & \textcolor{offgray}{69.11} &
\textcolor{offgray}{82.39} & \textcolor{offgray}{43.01} & \textcolor{offgray}{79.44} & \textcolor{offgray}{76.56} & \textcolor{offgray}{77.96} \\
\textcolor{offgray}{Llava-OV1.5-8B} &
\textcolor{offgray}{91.28} & \textcolor{offgray}{75.23} & \textcolor{offgray}{81.03} & \textcolor{offgray}{68.54} & \textcolor{offgray}{64.36} & \textcolor{offgray}{72.28} & \textcolor{offgray}{75.51} &
\textcolor{offgray}{48.48} & \textcolor{offgray}{46.62} & \textcolor{offgray}{40.86} & \textcolor{offgray}{45.80} &
\textcolor{offgray}{20.77} & \textcolor{offgray}{68.36} & \textcolor{offgray}{57.08} & \textcolor{offgray}{45.44} &
\textcolor{offgray}{82.33} & \textcolor{offgray}{79.84} & \textcolor{offgray}{74.50} & \textcolor{offgray}{75.00} & \textcolor{offgray}{87.79} & \textcolor{offgray}{73.17} &
\textcolor{offgray}{71.07} & \textcolor{offgray}{43.52} & \textcolor{offgray}{83.80} & \textcolor{offgray}{71.09} & \textcolor{offgray}{76.19} \\
\textcolor{offgray}{MiniCPM-V4.5-8B} &
\textcolor{offgray}{89.93} & \textcolor{offgray}{72.48} & \textcolor{offgray}{75.86} & \textcolor{offgray}{64.04} & \textcolor{offgray}{70.30} & \textcolor{offgray}{75.00} & \textcolor{offgray}{74.55} &
\textcolor{offgray}{48.82} & \textcolor{offgray}{54.05} & \textcolor{offgray}{63.98} & \textcolor{offgray}{54.52} &
\textcolor{offgray}{19.63} & \textcolor{offgray}{64.39} & \textcolor{offgray}{48.75} & \textcolor{offgray}{42.05} &
\textcolor{offgray}{87.07} & \textcolor{offgray}{80.93} & \textcolor{offgray}{73.37} & \textcolor{offgray}{76.85} & \textcolor{offgray}{87.46} & \textcolor{offgray}{70.33} &
\textcolor{offgray}{77.36} & \textcolor{offgray}{40.93} & \textcolor{offgray}{81.62} & \textcolor{offgray}{72.66} & \textcolor{offgray}{76.55} \\
VideoLLaMA3-7B &
90.60 & 68.81 & 45.22 & 31.03 & 20.21 & 68.16 & 54.00 &
48.82 & 60.81 & 27.42 & 45.32 &
22.35 & 69.95 & 51.25 & 45.88 &
77.29 & 79.29 & 71.39 & 53.70 & 77.23 & 51.22 & 64.15 & 31.09 & 82.87 & 66.41 & 68.90 \\
VideoChat-7B &
72.48 & 77.06 & 76.72 & 60.11 & 73.27 & 66.30 & 71.01 &
48.82 & 55.41 & 14.52 & 40.25 &
20.49 & 72.81 & 32.08 & 43.27 &
82.65 & 78.75 & 75.64 & 77.78 & 87.46 & 68.29 & 69.18 & 38.34 & 60.12 & 70.31 & 72.22 \\

\midrule
\multicolumn{27}{c}{\cellcolor{coffee!20}\textbf{Open-source Online Video-LLMs}} \\
\midrule
Flash-VStream-7B &
25.50 & 32.11 & 29.31 & 33.71 & 29.70 & 28.80 & 29.86 &
36.36 & 33.78 & 5.91 & 25.35 &
5.44 & 67.25 & 60.00 & 44.23 &
24.91 & 25.89 & 23.87 & 18.52 & 23.87 & 25.20 & 27.33 & 48.70 & 13.08 & 43.57 & 23.23 \\

Flash-VStream-7B* &
69.09 & 50.87 & 68.07 & 46.30 & 66.85 & 58.09 & 59.88 &
55.58 & 63.71 & 20.00 & 46.43 &
17.73 & 66.50 & 58.00 & 47.41 &
81.24 & 79.73 & 79.40 & 77.14 & 71.36 & 68.28 & 70.01 & 50.67 & 79.18 & 79.11 & 74.48 \\

ReKV-7B &
68.44 & 47.34 & 63.44 & 43.64 & 67.04 & 54.14 & 57.34 &
55.22 & 63.35 & 15.50 & 44.69 &
13.00 & 65.00 & 55.05 & 44.35 &
69.11 & 69.51 & 70.05 & 67.07 & 58.54 & 59.50 & 62.65 & 52.81 & 65.66 & 74.44 & 64.53 \\

StreamForest-7B &
68.46 & 53.21 & 71.55 & 47.75 & 65.35 & 60.87 & 61.20 &
58.92 & 64.86 & 32.26 & 52.02 &
32.81 & 70.59 & 57.08 & 53.49 &
82.65 & 83.11 & 84.26 & 76.85 & 75.64 & 69.11 & 77.50 & 54.40 & 78.19 & 82.81 & 77.26 \\
TimeChatOnline-7B &
69.80 & 48.60 & 64.70 & 44.90 & 68.30 & 55.40 & 58.60 &
53.90 & 62.80 & 9.10 & 42.00 &
32.50 & 36.50 & 40.00 & 36.40 &
78.86 & 79.13 & 80.77 & 77.78 & 66.19 & 67.07 & 70.44 & 53.72 & 77.26 & 81.25 & 73.64 \\

VideoChatOnline-4B &
47.22 & 40.21 & 46.68 & 39.20 & 48.65 & 41.86 & 43.97 &
49.05 & 54.78 & 8.37 & 37.40 &
19.07 & 54.00 & 46.00 & 39.69 &
62.99 & 63.47 & 63.56 & 60.35 & 53.74 & 54.75 & 57.76 & 52.24 & 58.38 & 70.17 & 58.81 \\

\bottomrule
\end{tabular}
\end{adjustbox}
\caption{Detailed experimental results for the experiments reported in Section 4.3}
\label{tab3}
\end{table*}

\paragraph{Experiment Details}
We conducted all experiments on GPUs provided by a cloud server. Notably, the GPU model and its performance characteristics do not correspond to mainstream, widely used GPU models. Therefore, to ensure transparency and reproducibility, we report in \ref{tab:perf-factors} the performance specifications provided by the cloud service vendor.We use compatible versions of Transformers and PyTorch, and enable FlashAttention-2 and Accelerate for runtime acceleration.Empirically, the overhead introduced by inter-process communication is negligible, and we observe no statistically significant degradation in end-to-end inference latency.

\paragraph{Experiment Setups.}
\noindent
To ensure resource comparability across multimodal models in an online inference setting, we adopt a unified, memory-footprint--based criterion derived from the online model to configure each model's context window. Specifically, we abstract the storage terms that scale approximately linearly with context length into two components: (i) multimodal representations written to the memory bank (e.g., embeddings of visual/audio tokens), and (ii) the Transformer attention cache (KV cache) associated with these multimodal tokens. Given a shared memory budget $M$ (in bytes), we use the online memory estimation function to compute each model's incremental storage cost per unit of context and then back-solve for the maximum admissible context length $B_i$ under budget $M$. Accordingly, we assign each multimodal model a context window that more closely matches its online memory consumption, enabling fairer and more reproducible cross-model comparisons without relying on raw token counts.

\paragraph{Experiment Results.}
To ensure consistency and comparability in the main results, we adopt a uniform weighting scheme in the main table, i.e., setting all weight coefficients equally as $w_f = w_a = w_t = w_r = 0.25$, and report the corresponding StreamingScore under this setting. We note that this configuration is intended as a default comprehensive evaluation protocol, and does not imply that all real-world deployment scenarios place identical emphasis on throughput, accuracy, time-to-first-token, and resource consumption. In practical applications, different users or tasks often prioritize different system capabilities, such as answer quality, interaction responsiveness, resource efficiency, or throughput. Therefore, assigning larger weights to the metrics that better align with specific deployment preferences can more accurately reflect the practical utility of a model in a given scenario. Based on this consideration, we further report results under different weight configurations in the appendix to demonstrate the applicability and robustness of StreamingScore across diverse deployment preferences.
To validate practical utility, we evaluate four representative deployment scenarios by setting the target term’s weight to 0.4 and the remaining three to 0.2. The table \ref{tab:scenario_ranking} below shows how representative models’ rankings change across scenarios.This scenario-aware metric captures strengths under distinct deployment constraints (e.g., the offline model \textbf{Qwen3-VL} becomes \textbf{Rank 1} in the ``best-answer'' setting, while the online model \textbf{Flash-VStream} is consistently strongest in ``interaction-first,'' ``resource-saving,'' and ``throughput-first'' settings). Importantly, despite scenario-specific re-ordering among top models, the overall ranking trend remains statistically robust across weight configurations (Spearman $\rho \in [0.972, 0.993]$, Kendall $\tau \in [0.909, 0.970]$), which reduces the risk that models with clear holistic weaknesses can game the leaderboard by merely tilting weights. 
\begin{table}[t]
\centering
\caption{Ranking changes of representative models under different deployment preferences.}
\label{tab:scenario_ranking}
\setlength{\tabcolsep}{4pt}
\renewcommand{\arraystretch}{1.1}
\resizebox{\linewidth}{!}{%
\begin{tabular}{lcccc}
\toprule
\textbf{Model / Type} &
\makecell[c]{\textbf{Best Answer} \\ ($w_a = 0.4$)} &
\makecell[c]{\textbf{Interaction First} \\ ($w_t = 0.4$)} &
\makecell[c]{\textbf{Edge Resource-Saving} \\ ($w_r = 0.4$)} &
\makecell[c]{\textbf{Throughput First} \\ ($w_f = 0.4$)} \\
\midrule
Qwen3-VL-7B (Offline)   & Rank 1 & Rank 2  & Rank 2  & Rank 2 \\
Flash-VStream (Online) & Rank 2 & Rank 1  & Rank 1  & Rank 1 \\
MiniCPM-V4.5 (Offline) & Rank 3 & Rank 3  & Rank 3  & Rank 5 \\
StreamForest (Online)  & Rank 9 & Rank 10 & Rank 9  & Rank 9 \\
\bottomrule
\end{tabular}
}
\end{table}

\paragraph{In-depth Discussion.}
As shown in Table~\ref{tab:ovo_detail}, different methods exhibit clear trade-offs across task clusters on OVO-Bench. Overall, open-source offline Video-LLMs are particularly strong on the ``Real-Time Visual Perception'' subset: Qwen3-VL achieves the highest average score in this subset (78.73), while InternVL3.5, LLaVA-OneVision, and MiniCTM remain at a comparable level (around 74--76), suggesting that the offline inference paradigm is more robust for fine-grained perception tasks (e.g., OCR, attribute recognition, and state understanding). In contrast, open-source online (streaming/online) Video-LLMs are generally weaker on the same perception subset, indicating that under online settings—constrained by temporal input processing and immediate generation requirements—low-level visual perception is more likely to become a bottleneck. Overall, offline methods tend to be stronger in fine-grained visual perception and robustness, while online methods need further improvements to close the gap in perceptual capability.

\paragraph{Notable Points about E2E Latency.}
Beyond the efficiency metrics reported in the main text, we conduct an additional analysis of end-to-end (E2E) latency. In streaming generation, E2E latency is directly tied to user experience: perceived waiting time is determined not only by the time-to-first-token (TTFT), but also by the efficiency of subsequent decoding and whether the model’s responses are concise and clear. We define E2E latency as the wall-clock time from query arrival to the completion of the final output, i.e., when the last token is generated and returned.

To measure this metric, we convert the OVO-Bench multiple-choice QA format into an open-ended generation setting and record E2E latency under a unified inference configuration. Importantly, the main-table results are obtained under a multiple-choice setup, whereas E2E latency can only be meaningfully measured in an open-ended generation scenario. Because these two evaluation settings differ fundamentally in output form, reporting E2E alongside the multiple-choice metrics in the main table could mislead readers into assuming they are directly comparable under an identical setup. For clarity and to avoid ambiguity, we therefore present the E2E experiments and discussions in the appendix.

\begin{figure}[t]
  \centering
  \includegraphics[width=\linewidth]{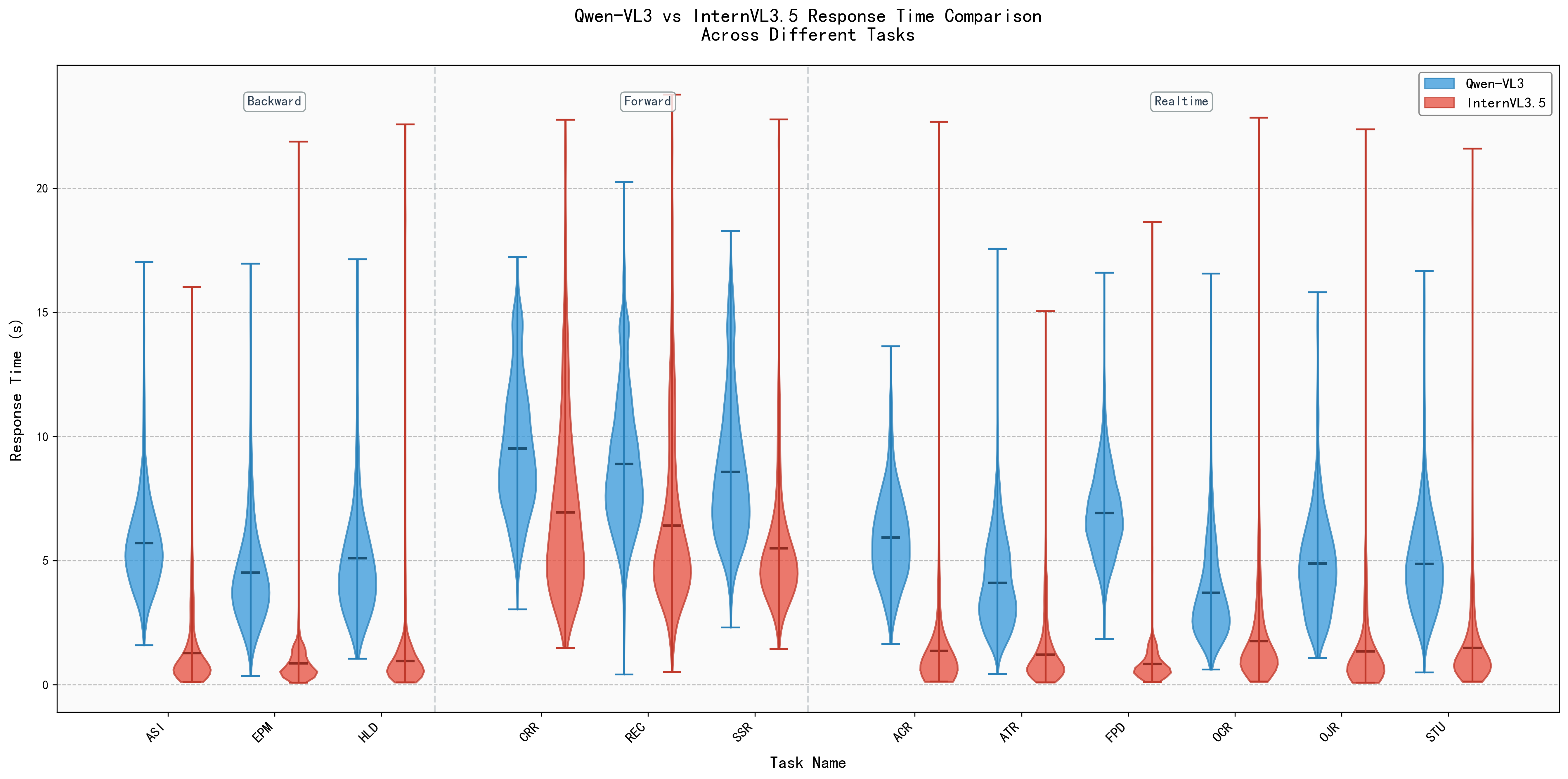}
  \caption{Comparison of two classic paradigms for video understanding.
}
  \label{compara}
\end{figure}

Under the same prompting and decoding configurations, E2E latency is largely influenced by the base language model’s generation style and stopping behavior; meanwhile, most models compared in the main text share similar base language models. Accordingly, in the appendix we focus on two representative families—Qwen3-VL and InternVL3.5—and provide a comparative analysis of their E2E latency on OVO-Bench.
As shown in Fig.~\ref{compara}, InternVL3.5 (red) consistently achieves lower end-to-end (E2E) latency than Qwen-VL3 (blue) across OVO-Bench tasks, with a tighter distribution and fewer long-tail cases. This trend holds across the Backward (ASI/EPM/HLD) and Realtime (ACR/ATR/FPD/OCR/OJR/STU) categories, where InternVL3.5 exhibits markedly smaller median latency and improved stability. For the Forward tasks (CRR/REC/SSR), both models show increased latency and heavier tails, reflecting the higher uncertainty and longer generations in open-ended settings; nevertheless, InternVL3.5 remains faster in central tendency. Overall, Fig.~\ref{compara} suggests that InternVL3.5 provides not only lower average latency but also more stable user-perceived responsiveness, whereas Qwen-VL3 is more susceptible to long-tail slowdowns.

\begin{table}[!htbp]
\centering
\small
\setlength{\tabcolsep}{4pt}
\renewcommand{\arraystretch}{1.05}
\begin{tabularx}{\columnwidth}{@{}>{\raggedright\arraybackslash}p{0.34\columnwidth}>{\raggedright\arraybackslash}X@{}}
\toprule
Factor & Value \\
\midrule
GPU model            & NVIDIA GeForce RTX 4090-48G \\
Peak throughput      & 40.32 TFLOPS \\
GPU memory (VRAM)    & 48 GB \\
GPU memory bandwidth & 1008.10 GB/s \\
PCIe lanes           & 16 \\
PCIe bandwidth       & 31.50 GB/s \\
CPU model            & Intel Xeon Platinum 8570 \\
Available CPU cores  & 20 \\
Available RAM        & 48 GB \\
Disk model           & SAMSUNG MZ7LH480 / SAMSUNG MZ7LH480 \\
Disk bandwidth       & 330.08 MB/s \\
Free disk space      & 200 GB \\
\bottomrule
\end{tabularx}
\caption{Hardware factors affecting performance}
\label{tab:perf-factors}
\end{table}

\paragraph{Detailed Analysis across Different Memory Budget.}
As shown in Table~\ref{tab:ovo_detail}, the three models exhibit consistent strengths across the three OVO-Bench task clusters. For Real-Time Visual Perception (OCR/ACR/ATR/STU/FPD/OJR), Qwen3-VL delivers the strongest and most stable overall performance, reaching a peak average of 78.73 at 0.5G and maintaining a high level (above 77) at larger scales; Llava-OV1.5 is typically second best, while InternVL3.5 is comparatively weaker and shows a mild degradation at the largest scale (e.g., 71.21 at 1.5G). For Backward Tracing (EPM/ASI/HLD), Qwen3-VL consistently leads (approximately 50.9--53.1 in average), whereas InternVL3.5 lags behind, largely due to persistently low HLD scores across scales. In contrast, the ranking on Forward Active Responding (REC/SSR/CRR) is more dynamic: InternVL3.5 surpasses Qwen3-VL at several scales and remains consistently stronger on REC, while Llava-OV1.5 benefits more from scaling and achieves the best cluster average at 1.0G/1.5G (49.72/50.00). Overall, Qwen3-VL is more competitive on perception and backward-tracing tasks, InternVL3.5 is comparatively stronger on active responding, and Llava-OV1.5 exhibits the largest scaling gains on forward tasks.
\paragraph{Detailed Analysis across Different Resolutions.}
As shown in Table~\ref{tab6}, increasing the input resolution (224$\times$224 $\rightarrow$ 336$\times$336 $\rightarrow$ 448$\times$448) leads to a structurally different impact across task clusters, and the scaling benefits vary by model. For Real-Time Visual Perception, higher resolution consistently yields the largest and most stable gains: Qwen3-VL improves from 67.98 to 73.36 and further to 75.75 in average score, while InternVL3.5 shows an even larger increase from 61.65 to 68.70 and then to 74.31, indicating that fine-grained visual cues (e.g., text details and local attributes) substantially benefit from higher-resolution inputs. In contrast, the effect on Backward Tracing is modest and less consistent: Qwen3-VL only slightly improves (49.76 $\rightarrow$ 50.55 $\rightarrow$ 51.35), and InternVL3.5 remains nearly flat (41.68 $\rightarrow$ 40.73 $\rightarrow$ 40.89), suggesting that the bottleneck of backward tracing is less dominated by visual detail. For Forward Active Responding, resolution scaling comdoes not provide a uniform advantage and can even be neutral or negative (e.g., Qwen3-VL slightly drops from 44.35 to 43.01 and 42.18), implying that these tasks are more sensitive to language-side generation and decision behaviors than to input resolution. Overall, resolution scaling primarily benefits perception-oriented tasks, while its impact on backward and forward tasks is comparatively limited and model-dependent.

\begin{table*}[t]
\centering
\scriptsize
\setlength{\tabcolsep}{3.2pt}
\renewcommand{\arraystretch}{1.15}
\begin{adjustbox}{max width=\textwidth,center}
\begin{tabular}{l|rrrrrrr|rrrr|rrrr}
\toprule
\multirow{3}{*}{\textbf{Model}} &
\multicolumn{15}{c}{\textbf{OVO-Bench}} \\
\cmidrule(lr){2-16}
& \multicolumn{7}{c|}{\textbf{Real-Time Visual Perception}} &
  \multicolumn{4}{c|}{\textbf{Backward Tracing}} &
  \multicolumn{4}{c}{\textbf{Forward Active Responding}} \\
\cmidrule(lr){2-8}\cmidrule(lr){9-12}\cmidrule(lr){13-16}
& OCR & ACR & ATR & STU & FPD & OJR & Avg. &
  EPM & ASI & HLD & Avg. &
  REC & SSR & CRR & Avg. \\
\midrule\midrule

\multicolumn{16}{c}{\textbf{0.1G}} \\
\midrule
\textcolor{offgray}{Qwen3-VL} &
\textcolor{offgray}{84.56} & \textcolor{offgray}{73.39} & \textcolor{offgray}{76.72} & \textcolor{offgray}{62.36} & \textcolor{offgray}{70.30} & \textcolor{offgray}{75.00} & \textcolor{offgray}{73.48} &
\textcolor{offgray}{42.76} & \textcolor{offgray}{54.73} & \textcolor{offgray}{60.75} & \textcolor{offgray}{50.87} &
\textcolor{offgray}{14.90} & \textcolor{offgray}{59.30} & \textcolor{offgray}{43.33} & \textcolor{offgray}{37.08} \\
\textcolor{offgray}{InternVL3.5} &
\textcolor{offgray}{87.25} & \textcolor{offgray}{76.15} & \textcolor{offgray}{75.86} & \textcolor{offgray}{66.29} & \textcolor{offgray}{69.31} & \textcolor{offgray}{70.65} & \textcolor{offgray}{73.95} &
\textcolor{offgray}{49.16} & \textcolor{offgray}{56.08} & \textcolor{offgray}{15.59} & \textcolor{offgray}{40.89} &
\textcolor{offgray}{19.77} & \textcolor{offgray}{57.07} & \textcolor{offgray}{45.42} & \textcolor{offgray}{38.67} \\
\textcolor{offgray}{Llava-OV1.5} &
\textcolor{offgray}{86.91} & \textcolor{offgray}{72.94} & \textcolor{offgray}{78.87} & \textcolor{offgray}{67.98} & \textcolor{offgray}{61.89} & \textcolor{offgray}{67.66} & \textcolor{offgray}{72.71} &
\textcolor{offgray}{44.44} & \textcolor{offgray}{42.90} & \textcolor{offgray}{48.66} & \textcolor{offgray}{45.33} &
\textcolor{offgray}{11.87} & \textcolor{offgray}{57.58} & \textcolor{offgray}{48.58} & \textcolor{offgray}{39.34} \\
\midrule

\multicolumn{16}{c}{\textbf{0.3G}} \\
\midrule
\textcolor{offgray}{Qwen3-VL} &
\textcolor{offgray}{92.62} & \textcolor{offgray}{77.98} & \textcolor{offgray}{77.59} & \textcolor{offgray}{69.66} & \textcolor{offgray}{70.30} & \textcolor{offgray}{76.63} & \textcolor{offgray}{77.54} &
\textcolor{offgray}{45.45} & \textcolor{offgray}{57.43} & \textcolor{offgray}{47.31} & \textcolor{offgray}{48.81} &
\textcolor{offgray}{19.63} & \textcolor{offgray}{64.07} & \textcolor{offgray}{47.50} & \textcolor{offgray}{41.74} \\
\textcolor{offgray}{InternVL3.5} &
\textcolor{offgray}{87.92} & \textcolor{offgray}{77.06} & \textcolor{offgray}{73.28} & \textcolor{offgray}{60.11} & \textcolor{offgray}{69.31} & \textcolor{offgray}{73.91} & \textcolor{offgray}{73.24} &
\textcolor{offgray}{50.84} & \textcolor{offgray}{56.08} & \textcolor{offgray}{13.44} & \textcolor{offgray}{41.05} &
\textcolor{offgray}{23.07} & \textcolor{offgray}{67.57} & \textcolor{offgray}{49.58} & \textcolor{offgray}{44.99} \\
\textcolor{offgray}{Llava-OV1.5} &
\textcolor{offgray}{91.28} & \textcolor{offgray}{75.69} & \textcolor{offgray}{78.02} & \textcolor{offgray}{68.54} & \textcolor{offgray}{61.89} & \textcolor{offgray}{70.10} & \textcolor{offgray}{74.25} &
\textcolor{offgray}{46.62} & \textcolor{offgray}{44.25} & \textcolor{offgray}{40.86} & \textcolor{offgray}{43.91} &
\textcolor{offgray}{15.40} & \textcolor{offgray}{64.74} & \textcolor{offgray}{52.26} & \textcolor{offgray}{44.14} \\
\midrule

\multicolumn{16}{c}{\textbf{0.5G}} \\
\midrule
\textcolor{offgray}{Qwen3-VL} &
\textcolor{offgray}{91.95} & \textcolor{offgray}{78.90} & \textcolor{offgray}{79.31} & \textcolor{offgray}{67.98} & \textcolor{offgray}{73.27} & \textcolor{offgray}{80.98} & \textcolor{offgray}{78.73} &
\textcolor{offgray}{48.82} & \textcolor{offgray}{61.49} & \textcolor{offgray}{48.92} & \textcolor{offgray}{51.82} &
\textcolor{offgray}{21.78} & \textcolor{offgray}{64.86} & \textcolor{offgray}{50.42} & \textcolor{offgray}{43.46} \\
\textcolor{offgray}{InternVL3.5} &
\textcolor{offgray}{88.59} & \textcolor{offgray}{75.23} & \textcolor{offgray}{77.59} & \textcolor{offgray}{61.80} & \textcolor{offgray}{71.29} & \textcolor{offgray}{73.91} & \textcolor{offgray}{74.31} &
\textcolor{offgray}{51.18} & \textcolor{offgray}{56.76} & \textcolor{offgray}{11.83} & \textcolor{offgray}{40.89} &
\textcolor{offgray}{25.36} & \textcolor{offgray}{67.73} & \textcolor{offgray}{50.00} & \textcolor{offgray}{46.14} \\
\textcolor{offgray}{Llava-OV1.5} &
\textcolor{offgray}{91.28} & \textcolor{offgray}{75.23} & \textcolor{offgray}{81.03} & \textcolor{offgray}{68.54} & \textcolor{offgray}{64.36} & \textcolor{offgray}{72.28} & \textcolor{offgray}{75.51} &
\textcolor{offgray}{48.48} & \textcolor{offgray}{46.62} & \textcolor{offgray}{40.86} & \textcolor{offgray}{45.80} &
\textcolor{offgray}{20.77} & \textcolor{offgray}{68.36} & \textcolor{offgray}{57.08} & \textcolor{offgray}{45.44} \\
\midrule

\multicolumn{16}{c}{\textbf{1.0G}} \\
\midrule
\textcolor{offgray}{Qwen3-VL} &
\textcolor{offgray}{91.95} & \textcolor{offgray}{76.15} & \textcolor{offgray}{80.17} & \textcolor{offgray}{66.29} & \textcolor{offgray}{71.29} & \textcolor{offgray}{78.80} & \textcolor{offgray}{77.42} &
\textcolor{offgray}{50.84} & \textcolor{offgray}{64.86} & \textcolor{offgray}{47.31} & \textcolor{offgray}{53.09} &
\textcolor{offgray}{24.64} & \textcolor{offgray}{67.41} & \textcolor{offgray}{52.92} & \textcolor{offgray}{46.14} \\
\textcolor{offgray}{InternVL3.5} &
\textcolor{offgray}{87.25} & \textcolor{offgray}{72.48} & \textcolor{offgray}{75.86} & \textcolor{offgray}{60.67} & \textcolor{offgray}{71.29} & \textcolor{offgray}{72.28} & \textcolor{offgray}{72.88} &
\textcolor{offgray}{51.18} & \textcolor{offgray}{57.43} & \textcolor{offgray}{10.22} & \textcolor{offgray}{40.57} &
\textcolor{offgray}{29.37} & \textcolor{offgray}{66.61} & \textcolor{offgray}{54.58} & \textcolor{offgray}{48.18} \\
\textcolor{offgray}{Llava-OV1.5} &
\textcolor{offgray}{90.61} & \textcolor{offgray}{72.48} & \textcolor{offgray}{80.60} & \textcolor{offgray}{67.13} & \textcolor{offgray}{63.37} & \textcolor{offgray}{70.38} & \textcolor{offgray}{74.09} &
\textcolor{offgray}{49.49} & \textcolor{offgray}{48.64} & \textcolor{offgray}{39.25} & \textcolor{offgray}{45.79} &
\textcolor{offgray}{22.62} & \textcolor{offgray}{67.49} & \textcolor{offgray}{59.04} & \textcolor{offgray}{49.72} \\
\midrule

\multicolumn{16}{c}{\textbf{1.5G}} \\
\midrule
\textcolor{offgray}{Qwen3-VL} &
\textcolor{offgray}{91.28} & \textcolor{offgray}{77.06} & \textcolor{offgray}{79.31} & \textcolor{offgray}{67.98} & \textcolor{offgray}{74.26} & \textcolor{offgray}{77.17} & \textcolor{offgray}{77.66} &
\textcolor{offgray}{51.85} & \textcolor{offgray}{65.54} & \textcolor{offgray}{43.01} & \textcolor{offgray}{52.46} &
\textcolor{offgray}{25.93} & \textcolor{offgray}{67.73} & \textcolor{offgray}{53.33} & \textcolor{offgray}{46.90} \\
\textcolor{offgray}{InternVL3.5} &
\textcolor{offgray}{85.91} & \textcolor{offgray}{69.72} & \textcolor{offgray}{75.86} & \textcolor{offgray}{60.11} & \textcolor{offgray}{73.27} & \textcolor{offgray}{66.85} & \textcolor{offgray}{71.21} &
\textcolor{offgray}{56.57} & \textcolor{offgray}{61.49} & \textcolor{offgray}{11.83} & \textcolor{offgray}{44.53} &
\textcolor{offgray}{31.09} & \textcolor{offgray}{63.75} & \textcolor{offgray}{53.75} & \textcolor{offgray}{47.67} \\
\textcolor{offgray}{Llava-OV1.5} &
\textcolor{offgray}{89.60} & \textcolor{offgray}{71.56} & \textcolor{offgray}{80.16} & \textcolor{offgray}{67.70} & \textcolor{offgray}{65.85} & \textcolor{offgray}{66.85} & \textcolor{offgray}{73.62} &
\textcolor{offgray}{52.69} & \textcolor{offgray}{51.01} & \textcolor{offgray}{37.91} & \textcolor{offgray}{47.20} &
\textcolor{offgray}{24.40} & \textcolor{offgray}{66.50} & \textcolor{offgray}{59.10} & \textcolor{offgray}{50.00} \\
\bottomrule
\end{tabular}
\end{adjustbox}
\caption{Detailed experimental results for the experiments reported in Section 4.4}
\label{tab:ovo_detail}
\end{table*}

\begin{table*}[t]
\centering
\scriptsize
\setlength{\tabcolsep}{3.2pt}
\renewcommand{\arraystretch}{1.15}
\begin{adjustbox}{max width=\textwidth,center}
\begin{tabular}{l|rrrrrrr|rrrr|rrrr}
\toprule
\multirow{3}{*}{\textbf{Model}} &
\multicolumn{15}{c}{\textbf{OVO-Bench}} \\
\cmidrule(lr){2-16}
& \multicolumn{7}{c|}{\textbf{Real-Time Visual Perception}} &
  \multicolumn{4}{c|}{\textbf{Backward Tracing}} &
  \multicolumn{4}{c}{\textbf{Forward Active Responding}} \\
\cmidrule(lr){2-8}\cmidrule(lr){9-12}\cmidrule(lr){13-16}
& OCR & ACR & ATR & STU & FPD & OJR & Avg. &
  EPM & ASI & HLD & Avg. &
  REC & SSR & CRR & Avg. \\
\midrule\midrule

\multicolumn{16}{c}{\textbf{224*224}} \\
\midrule
\textcolor{offgray}{Qwen3-VL} &
\textcolor{offgray}{69.80} & \textcolor{offgray}{68.81} & \textcolor{offgray}{76.72} & \textcolor{offgray}{59.55} & \textcolor{offgray}{67.33} & \textcolor{offgray}{69.02} & \textcolor{offgray}{67.98} &
\textcolor{offgray}{50.51} & \textcolor{offgray}{60.81} & \textcolor{offgray}{39.78} & \textcolor{offgray}{49.76} &
\textcolor{offgray}{22.78} & \textcolor{offgray}{67.25} & \textcolor{offgray}{47.08} & \textcolor{offgray}{44.35} \\
\textcolor{offgray}{InternVL3.5} &
\textcolor{offgray}{65.77} & \textcolor{offgray}{59.63} & \textcolor{offgray}{66.38} & \textcolor{offgray}{58.99} & \textcolor{offgray}{62.38} & \textcolor{offgray}{58.70} & \textcolor{offgray}{61.65} &
\textcolor{offgray}{51.85} & \textcolor{offgray}{59.46} & \textcolor{offgray}{11.29} & \textcolor{offgray}{41.68} &
\textcolor{offgray}{24.64} & \textcolor{offgray}{62.16} & \textcolor{offgray}{59.58} & \textcolor{offgray}{45.05} \\
\textcolor{offgray}{FlashVStream-LLaVA} &
\textcolor{offgray}{25.50} & \textcolor{offgray}{32.11} & \textcolor{offgray}{29.31} & \textcolor{offgray}{33.71} & \textcolor{offgray}{29.70} & \textcolor{offgray}{28.80} & \textcolor{offgray}{29.86} &
\textcolor{offgray}{36.36} & \textcolor{offgray}{33.78} & \textcolor{offgray}{5.91} & \textcolor{offgray}{25.35} &
\textcolor{offgray}{5.44} & \textcolor{offgray}{67.25} & \textcolor{offgray}{60.00} & \textcolor{offgray}{44.23} \\
\midrule

\multicolumn{16}{c}{\textbf{336*336}} \\
\midrule
\textcolor{offgray}{Qwen3-VL} &
\textcolor{offgray}{81.88} & \textcolor{offgray}{78.90} & \textcolor{offgray}{76.72} & \textcolor{offgray}{64.61} & \textcolor{offgray}{70.30} & \textcolor{offgray}{71.20} & \textcolor{offgray}{73.36} &
\textcolor{offgray}{49.83} & \textcolor{offgray}{59.46} & \textcolor{offgray}{44.62} & \textcolor{offgray}{50.55} &
\textcolor{offgray}{20.92} & \textcolor{offgray}{66.45} & \textcolor{offgray}{45.83} & \textcolor{offgray}{43.01} \\
\textcolor{offgray}{InternVL3.5} &
\textcolor{offgray}{79.19} & \textcolor{offgray}{67.89} & \textcolor{offgray}{71.55} & \textcolor{offgray}{58.43} & \textcolor{offgray}{72.28} & \textcolor{offgray}{66.85} & \textcolor{offgray}{68.70} &
\textcolor{offgray}{52.53} & \textcolor{offgray}{56.76} & \textcolor{offgray}{9.14} & \textcolor{offgray}{40.73} &
\textcolor{offgray}{27.51} & \textcolor{offgray}{65.02} & \textcolor{offgray}{56.67} & \textcolor{offgray}{47.03} \\
\textcolor{offgray}{FlashVStream-LLaVA} &
\textcolor{offgray}{30.31} & \textcolor{offgray}{36.69} & \textcolor{offgray}{30.45} & \textcolor{offgray}{34.98} & \textcolor{offgray}{32.71} & \textcolor{offgray}{31.25} & \textcolor{offgray}{32.73} &
\textcolor{offgray}{36.35} & \textcolor{offgray}{32.64} & \textcolor{offgray}{5.71} & \textcolor{offgray}{24.90} &
\textcolor{offgray}{5.53} & \textcolor{offgray}{68.40} & \textcolor{offgray}{57.74} & \textcolor{offgray}{43.89} \\
\midrule

\multicolumn{16}{c}{\textbf{448*448}} \\
\midrule
\textcolor{offgray}{Qwen3-VL} &
\textcolor{offgray}{86.58} & \textcolor{offgray}{77.06} & \textcolor{offgray}{75.00} & \textcolor{offgray}{66.29} & \textcolor{offgray}{73.27} & \textcolor{offgray}{77.17} & \textcolor{offgray}{75.75} &
\textcolor{offgray}{51.52} & \textcolor{offgray}{58.11} & \textcolor{offgray}{45.70} & \textcolor{offgray}{51.35} &
\textcolor{offgray}{19.34} & \textcolor{offgray}{66.45} & \textcolor{offgray}{45.00} & \textcolor{offgray}{42.18} \\
\textcolor{offgray}{InternVL3.5} &
\textcolor{offgray}{88.59} & \textcolor{offgray}{75.23} & \textcolor{offgray}{77.59} & \textcolor{offgray}{61.80} & \textcolor{offgray}{71.29} & \textcolor{offgray}{73.91} & \textcolor{offgray}{74.31} &
\textcolor{offgray}{51.18} & \textcolor{offgray}{56.76} & \textcolor{offgray}{11.83} & \textcolor{offgray}{40.89} &
\textcolor{offgray}{25.36} & \textcolor{offgray}{67.73} & \textcolor{offgray}{50.00} & \textcolor{offgray}{46.14} \\
\textcolor{offgray}{FlashVStream-LLaVA} &
\textcolor{offgray}{32.99} & \textcolor{offgray}{38.24} & \textcolor{offgray}{31.46} & \textcolor{offgray}{36.42} & \textcolor{offgray}{33.13} & \textcolor{offgray}{34.23} & \textcolor{offgray}{34.41} &
\textcolor{offgray}{36.49} & \textcolor{offgray}{32.26} & \textcolor{offgray}{6.49} & \textcolor{offgray}{25.08} &
\textcolor{offgray}{5.11} & \textcolor{offgray}{69.86} & \textcolor{offgray}{53.85} & \textcolor{offgray}{42.94} \\
\bottomrule
\end{tabular}
\end{adjustbox}
\caption{Detailed experimental results for the experiments reported in Section 4.4}
\label{tab6}
\end{table*}

\end{document}